\documentclass{ecai}
\usepackage{times}
\usepackage{graphicx}
\usepackage{latexsym}
\usepackage{pgfplots}
\pgfplotsset{width=10cm,compat=1.10}
\usepackage{float}
\restylefloat{table}

\usepackage{amsmath,amsthm,amssymb}
\usepackage{xcolor}

\usepackage{url}

\begin{document}

\title{Entity Embeddings with Conceptual Subspaces as a Basis for Plausible Reasoning}

\author{Shoaib Jameel\institute{Cardiff University,
UK, email: JameelS1@cardiff.ac.uk} \and Steven Schockaert\institute{Cardiff University,
UK, email: SchockaertS1@cardiff.ac.uk} }

\maketitle

\begin{abstract}
Conceptual spaces are geometric representations of conceptual knowledge in which entities correspond to points, natural properties correspond to convex regions, and the dimensions of the space correspond to salient features. While conceptual spaces enable elegant models of various cognitive phenomena, the lack of automated methods for constructing such representations have so far limited their application in artificial intelligence. To address this issue, we propose a method which learns a vector-space embedding of entities from Wikipedia and constrains this embedding such that entities of the same semantic type are located in some lower-dimensional subspace. We experimentally demonstrate the usefulness of these subspaces as approximate conceptual space representations by showing, among others, that important features can be modelled as directions and that natural properties tend to correspond to convex regions.
 \end{abstract}

\section{INTRODUCTION}
Despite the fact that several large-scale open-domain knowledge bases are now available 
(e.g.\ CYC, SUMO, Freebase, Wikidata and YAGO), 
few knowledge-driven applications rely on logical reasoning. An important reason for this is that available knowledge is often inconsistent. For example, the concept \textit{ice cream shop} is asserted to be disjoint from \textit{restaurant} in CYC, while it is considered a type of \textit{restaurant} on Wikipedia\footnote{\url{https://en.wikipedia.org/wiki/Category:Types_of_restaurants}}.  Another challenge for logical reasoning is that available knowledge is seldom complete. For example, SUMO encodes\footnote{\url{https://github.com/ontologyportal/sumo/blob/master/Sports.kif}} knowledge about chess, darts and poker, but mentions nothing about checkers.

Humans are remarkably adept at overcoming such challenges \cite{collins1989logic,festinger1962theory}. For example, we can recognize that the aforementioned conflict between CYC and Freebase is caused by the vagueness of the categories \textit{restaurant} and \textit{shop}, which both have \textit{ice cream shop} as a borderline case. Similarly, we can deal with knowledge gaps by making inductive inferences, e.g.\ assuming that properties which hold for chess, darts and poker should hold for checkers as well. 
Automating such forms of plausible reasoning has proven challenging, among others because they rely on an underlying notion of similarity, which is difficult to characterize using purely symbolic methods.

The solution offered by the theory of conceptual spaces \cite{Gardenfors:conceptualSpaces} is to represent concepts as regions in a suitable metric space. The points of this space correspond to (actual or possible) entities of a given semantic type, such that similar entities are located close to each other. It is furthermore posited that most natural properties correspond to convex regions, in accordance with prototype theory \cite{rosch1973natural}. Furthermore, the dimensions of a conceptual space correspond to the salient features of the considered domain. For example, a conceptual space of wines could have dimensions relating to sweetness, acidity, fruitiness, amount of tannins, etc. 
Using conceptual space representations, many cognitive phenomena, including vagueness and induction, can be modelled in a natural way \cite{Gardenfors:conceptualSpaces,douven2013vagueness,schockaert2013interpolative,DBLP:journals/connection/LietoMPR15}. However, existing applications have focused on a few particular domains in which conceptual space representations can be derived from available metric information. For example, several authors have considered conceptual spaces for music perception \cite{forth2010unifying,musicConceptualSpacesChella}. In such cases, the definition of the conceptual space, and its relationship to e.g.\ audio signals, relies on well-understood insights from the field of music cognition.

The research question we consider in this paper is whether we can automatically obtain approximate conceptual space representations for a wide range of domains, by combining information found in existing knowledge bases with representations derived from large text corpora such as Wikipedia. 

Our approach builds on existing work for learning word embeddings from text corpora. Similar to conceptual spaces, word embeddings \cite{DBLP:conf/nips/MikolovSCCD13,glove2014,Turney10} represent the meaning of words in a high-dimensional Euclidean space, typically as vectors\footnote{Two notable exceptions are \cite{Erk:2009:RWR:1596374.1596387} and \cite{DBLP:journals/corr/VilnisM14}, where words are represented using densities}. There are, however, two important differences between word embeddings and conceptual spaces. First, while word embeddings represent all words in a single vector space, conceptual spaces model the entities (and their properties) of a particular semantic type only (e.g.\ people and cities would be modelled in separate conceptual spaces). Because of this restriction, conceptual spaces can have dimensions that reflect the salient properties of the underlying domain. This allows us to use conceptual spaces for ranking entities (e.g.\ a conceptual space of cities should have a dimension corresponding the population, allowing us to rank cities from the least to the most populous), modelling context effects\footnote{The context-dependent nature of similarity is modelled in conceptual spaces by allowing dimensions to be rescaled, depending on the importance of the corresponding property in the given context.}, and for describing how two entities or concepts are semantically related (e.g.\ that the rules of chess are more complex than the rules of checkers). In contrast, the dimensions of a word embedding space are essentially meaningless. Second, conceptual spaces clearly differentiate entities, which are modelled as points, from properties, which are modelled as regions. As a result, conceptual spaces can be used to model that a given entity has a given property or belongs to a given category, to model typicality (e.g.\ an \textit{ice cream shop} could be located in the region modelling \textit{shop} but towards the border), and to model semantic relations between different properties and categories. 

In \cite{derracAIJ} an approach was proposed for learning conceptual space representations, which consists of (i) representing each entity of a given semantic type as a bag of words (e.g.\ each movie is represented as its set of user reviews), (ii) converting that bag of words representation to a vector space representation using multi-dimensional scaling (MDS), and (iii) identifying directions corresponding to salient properties of the considered domain in a post-hoc analysis.

An important limitation of the approach from \cite{derracAIJ} is that it cannot take advantage of relationships between different conceptual spaces. To address this, in this paper we propose a method that learns a single domain-independent vector space, in which each semantic type corresponds to a particular subspace. In other words, we learn conceptual space representations which are themselves embedded in a higher-dimensional vector space. Among others, this allows us to model semantic type hierarchies (e.g.\ the conceptual space of humans, in our model, is a subspace of the conceptual space of living things). Furthermore, different conceptual spaces can be aligned by taking into account semantic relations between entities of the corresponding types (e.g.\ the subspaces representing actors, directors and genres can help to obtain a more accurate representation of movies). Another important limitation of the approach from \cite{derracAIJ} is that MDS requires a distance matrix whose size is quadratic in the number of entities, which severely limits its scalability. In contrast, our model can easily learn representations for millions of entities.


\section{RELATED WORK}\label{secRelatedWork}
\subsection{Word embedding} 
Word embeddings are vector space representations which are used to model the meaning of words.
Several existing models construct a vector for each word by applying some form of matrix factorization to a term-term co-occurrence matrix; see \cite{Turney10} for an overview of such approaches. Recently, a number of models have been proposed which instead explicitly optimize the predictive power of the word vectors. For example, the popular Skip-gram model \cite{DBLP:conf/nips/MikolovSCCD13} tries to find word vectors that can be used to predict the probability of seeing a context word, given an occurrence of the word being modelled, while the related continuous bag-or-words (CBOW) model focuses on the probability of seeing the word being modelled, given the occurrence of a context word.

An interesting property of word embeddings is that they often capture several kinds of semantic relations, beyond simple similarity. For example, in \cite{DBLP:conf/nips/MikolovSCCD13} it is shown that analogical proportions of the form $a$ is to $b$ what $c$ is to $d$ correspond to approximate parallelograms in the  space obtained by Skip-gram. 
They also found that vector addition sometimes corresponds to a form of semantic composition, e.g.\ adding the vectors for \textit{Germany} and \textit{capital} resulted in a vector which is close to the vector for \textit{Berlin}.

The fact that the vector space obtained by the Skip-gram model satisfies such linear regularities is at first glance somewhat surprising. In \cite{glove2014}, the authors analyze what characteristics of a word embedding model can explain this effect, and propose a new model, called GloVe, which is explicitly aimed at capturing linear regularities. Since our model will build on GloVe, we briefly review its formulation. The GloVe model relies on a term-term co-occurrence matrix $X=(x_{ij})$, where $x_{ij}$ is the number of times that word \(i\) appears in the context of word \(j\). For each term $t_i$ in the vocabulary, two word vectors $w_i$ and $\tilde{w}_i$ and a  bias $b_i$ are chosen by minimizing the following objective:
\begin{equation}\label{eqDefGlove}
 J=\sum_{i=1}^{V} \sum_{j=1}^{V}f(x_{ij})(w_i \cdot \tilde{w}_j+b_i+b_j-\log x_{ij})^2
\end{equation}
where $V$ is the number of words in the vocabulary. The function $f$ is used to limit the impact of rare terms, whose co-occurrence counts are considered to be noisy. It is defined as follows:
\begin{align}\label{eqDefGloVeF}
 f(x_{ij}) = \begin{cases} 
                       \bigg( \frac{x_{ij}}{x_{\textit{max}}}\bigg)^{\alpha} & \text{ if } x_{ij} < x_{\text{max}} \\
                       1 & \text{ otherwise}
                      \end{cases}
\end{align}
where $x_{\textit{max}}$ is a constant which was fixed as 100.
Intuitively, $w_i$ reflects the meaning of term $t_i$ while $\tilde{w}_j$ reflects how the occurrence of that term in the context of another term $t_j$ impacts the meaning of $t_j$.


\subsection{Knowledge graph embedding} 
Knowledge bases such as Freebase and Wikidata can essentially be seen as collections of (subject, predicate, object) triples, and can thus be encoded as a graph, where nodes correspond to entities and edges are labelled with relation types. Several authors have looked at the problem of automatically expanding such knowledge graphs \cite{dong2014knowledge}. Here, we focus on models that rely on embedding knowledge graphs in a vector space, as we will use similar ideas for aligning different conceptual subspaces. The idea of embedding knowledge graphs in a vector space was proposed in \cite{BordesWCB11}. In particular, they propose the model SE, in which each entity $e_i$ is represented as a vector and each relation $r_k$ is represented using two matrices $R_k^{\textit{lhs}}$ and $R_k^{\textit{rhs}}$. The constraint they impose is that the following distance should be small for triples $(e_i,r_k,e_j)$ in the knowledge graph and large for other triples:
\begin{align*}
d( R_k^{\textit{lhs}} e_i,  R_k^{\textit{rhs}} e_j)
\end{align*}
where $d$ is either the Euclidean or Manhattan distance. An important drawback of this model is that it requires learning a large number of parameters, which was empirically found to lead to underfitting \cite{NIPS2013_5071}. In \cite{NIPS2013_5071} a simpler alternative, called TransE, was proposed, which represents each relation as a vector and considers the following scoring function instead:

\begin{align*}
d(e_i + r_k, e_j)
\end{align*}
Despite the simplicity of this model, it was shown to substantially outperform SE in practice. However, as noted in \cite{DBLP:conf/aaai/WangZFC14}, TransE is mostly suitable for one-to-one relations. To obtain a more faithful modelling of one-to-many, many-to-one and many-to-many relations, the model TransH is proposed. In this model, both a hyperplane $H_k$ and an $(n-1)$ dimensional vector $r_k$ is associated with each relation type (with $n$ the dimension of the embedding space), and the following scoring function is considered:
\begin{align*}
d(e_i^{H_k} + r_k, e_j^{H_k})
\end{align*}
where $e_i^{H_k}$ and $e_j^{H_k}$ are the orthogonal projections of $e_i$ and $e_j$ on the hyperplane $H_k$. The TransR model, introduced in \cite{DBLP:conf/aaai/LinLSLZ15}, follows a similar strategy, but instead associates an $m$-dimensional vector $r_k$ and an $m\times n$ matrix $M_k$ with each relation, and uses the following scoring function:

\begin{align*}
d(e_i M_r + r_k, e_j M_r)
\end{align*}
The underlying idea is to use the TransE model, after projecting the entities onto a relation-specific space. While in general it is not required that $n=m$, this particular choice was used in all experiments.  Finally, \cite{DBLP:conf/aaai/LinLSLZ15} also proposes a variant CTransR, in which each entities are clustered, and each relation can have a different representation for each cluster.

In our model, the semantic types of entities play a crucial role. One other approach that explicitly takes semantic type into account is \cite{DBLP:conf/acl/GuoWWWG15}. In particular, they add a regularization term to the objective function of existing embedding models to encode the requirement that entities of the same semantic type should be represented using similar vectors, which they formalize based on two manifold learning algorithms. Unfortunately, the scalability of the resulting method is relatively limited.

In \cite{wang2014knowledge} a model is proposed that combines word embedding with knowledge graph embedding. In particular, they jointly learn a representation for words, entities and relations, where the word representations are constrained similarly as in the Skip-gram model and the entities and relations are constrained similarly as in the TransE model. The entity and word representations are aligned either based on Wikipedia anchors or based on the entity names. An improvement of this method was proposed in \cite{DBLP:conf/emnlp/ZhongZWWC15}, where the alignment is instead based on the text of the Wikipedia article of the entity. Along similar lines, \cite{DBLP:conf/cikm/XuBBGWLL14} proposes a model in which the objective functions of Skip-gram and TransE are combined.  A third component in their objective function allows the model to take into account an external similarity relation, by imposing the requirement that similar terms should have similar vectors. It is shown that the resulting model improves the word embeddings from Skip-gram. 

The model we propose in this paper also combines a word based entity embedding component with a knowledge graph embedding component, although our motivation is different. In particular, \cite{wang2014knowledge} and \cite{DBLP:conf/emnlp/ZhongZWWC15} add a word based entity embedding component to a knowledge graph embedding model to improve the predictive performance for entities about which little or nothing is included in the knowledge graph. For example, if the knowledge graph contains the fact that entity $a$ is in relation $R$ with entity $b$, and from the word based component we can derive that entity $a'$ is similar to entity $a$, then we can plausibly derive that entity $a'$ might also be in relation $R$ with entity $b$. Intuitively, we can thus view these approaches as using word based entity embedding to add a kind of smoothing to the knowledge graph embedding model. In contrast, our aim is to model how different entities of the same type are related. The kind of semantic relatedness in which we are interested (e.g.\ modelling that one building is \textit{taller than} another one) is typically not captured by existing knowledge graphs. Intuitively, we use the word based entity embedding component to learn domain-specific vector space representations, and then use a knowledge graph embedding model to align these spaces. This allows us to improve the representation for semantic types about which little information is available in the considered textual descriptions. In this sense, we can view our model as using the knowledge graph embedding component to add a kind of smoothing to the word based entity embedding. Our motivation is somewhat similar in spirit to  \cite{DBLP:conf/cikm/XuBBGWLL14}, but that model focuses on word embeddings rather than entity embeddings.

To the best of our knowledge, our model is the first to use semantic type information to learn domain-specific subspaces.




\section{DESCRIPTION OF THE MODEL}\label{secModel}
Our aim is to learn a vector-space embedding of a set of entities $E$, in which entities of the same semantic type lie in some lower-dimensional subspace. Let $S$ be the set of all semantic types. For $s\in S$, we write $E_s$ for the set of all entities of type $s$. We furthermore assume that a set of binary relations $R$ is available,  and a set $G\subseteq E\times R \times E$ of triples of the form $(e,k,f)$, encoding that entities $e$ and $f$ are in relation $k$. Finally, we assume that for every entity e, a bag of words $W_e$ describing that entity is available. 
The model we propose has the following form:
\begin{align}\label{eqModelOverview}
J = \alpha J_{\textit{text}} + (1-\alpha)(J_{\textit{type}} +  J_{\textit{rel}}) + \beta J_{\textit{reg}}
\end{align}
where $\alpha\in[0,1]$ and $\beta\in [0,+\infty[$ are parameters controlling the relative importance of the different components of the model. Component $J_{\textit{text}}$ will be used to constrain the representation of the entities based on their textual description, $J_{\textit{type}}$ will impose the constraint that entities of the same type belong to a particular subspace, $J_{\textit{rel}}$ will use the relations in $R$ to improve the alignment between these subspaces, and $J_{\textit{reg}}$ is a regularization component which will allow the model to automatically select the most appropriate number of dimensions for every subspace. We now discuss each of these components in more detail.

\subsection{Word based entity embedding}
From the bag of words representations $W_e$, we want to find a point $p_e \in \mathbb{R}^n$ for each entity $e$ such that similar entities correspond to nearby points and such that salient features can be interpreted as directions in the space. Specifically, let $f$ be a feature of interest, and let $x_i \in \mathbb{R}$ be the value of feature $f$ for entity $e_i$, i.e.\ $x_i$ reflects how much $e_i$ has feature $f$. Then there should be a vector $w_f \in \mathbb{R}^n$ such that the orthogonal projection $p_{e_i}'$ of the point $p_{e_i}$ on the line $L_f = \{ q \,|\, q = \lambda \cdot w_f, \lambda\in\mathbb{R}\}$ is given by $p_{e_i}' = c_f x_i  w_f + b_f$ for $b_f$ and $c_f$ constants in $\mathbb{R}$. In other words, in a coordinate system where $L_f$ coincides with one of the axes, the corresponding coordinate of $p_e$ should be proportional to $x_i$. This requirement is equivalent to\footnote{We are abusing notation here, using $p_{e_i}$ as a notation for the vector $\overrightarrow{0p_{e_i}}$.}:
\begin{align}\label{eqConstraintFeatures1}
p_{e_i} \cdot w_f  = (c_f x_i + b_f) \cdot \|w_f\|
\end{align}
Unfortunately, we do not actually know what are the salient features in most domains. Following \cite{derracAIJ}, we therefore use word co-occurrence as a proxy for feature values. In particular, we assume that each word potentially corresponds to a salient feature, and that the number of times a word co-occurs with a given entity reflects how much that entity has the corresponding feature. This leads to the following constraint
\begin{align}\label{eqConstraintFeatures2}
p_{e_i} \cdot w_j  = g(y_{ji}) + b_j
\end{align}
where $y_{ji}$ is the number of times word $t_j$ occurs in $W_{e_i}$, $g$ is a monotonic function that maps co-occurrence statistics to feature values, and $b_j$ is a constant. Typically it will not be possible to satisfy the constraint \eqref{eqConstraintFeatures2} for all entities and all context words. The assumption underlying this model is that the salient features of an entity affect the co-occurrence statistics of many context words, and that the words for which \eqref{eqConstraintFeatures2} is (approximately) satisfied, in an optimal solution, will therefore be those that are strongly related to important features of the entity $e_i$.

Note that the requirement in \eqref{eqConstraintFeatures2} closely resembles the constraints that are optimized by the GloVe model. Moreover, as in the GloVe model, we can choose $g(y_{ji})=\log(y_{ji}) - b_i$ and formalize the objective function as a least squares regression problem, weighted such that frequent terms have a stronger impact on the objective function:
\begin{align*}
 J_{\textit{text}}^E=\sum_{e_i \in E} \sum_{t_j \in W_{e_i}} f(y_{ji})(p_{e_i} \cdot w_j+b_i+b_j-\log y_{ji})^2
\end{align*}
where $f$ is defined as in \eqref{eqDefGloVeF}.
The resulting model is essentially the same as GloVe, but instead of modelling word-word co-occurrence we now model entity-word co-occurrence. The geometric interpretation, however, is different, as we view entities as points and context words as vectors. We can further constrain the word vectors $w_j$ by adding a second component, capturing word-word co-occurrences, which corresponds to the original GloVe model. In particular, we define $ J_{\textit{text}} =  J_{\textit{text}}^E + J_{\textit{glove}}$, where $J_{\textit{glove}}$ is the objective function $J$ defined in \eqref{eqDefGlove}.

\subsection{Subspace constraints}
A key distinguishing feature of our model is that all entities of a given type $s$ are imposed to belong to the same subspace.
To formalize this constraint, we associate with each semantic type $s$ a set of $n+1$ points $p_0^s,...,p_{n}^s \in \mathbb{R}^n$ and express that for each entity $e_i$ of type $s$, the point $p_{e_i}$ can be written as a convex combination of the points $p_0^s,...,p_{n}^s$:
\begin{align*}
J_{\textit{type}} = \sum_{s\in S} \sum_{e\in E_s} \|p_{e} - \sum_{j=0}^{n} \lambda_{j}^{e,s} p_j^s\|^2 
\end{align*}
where we impose that $\lambda_j^{e,x}\geq 0$ and $\sum_{j=0}^{n} \lambda_{j}^{e,s} = 1$. Note that on its own, this component is trivial, as it suffices to choose any set of points $p_0^s,...,p_{n}^s$ in general linear position. However, we will additionally require that the space spanned by the points $p_0^s,...,p_{n}^s$ is as low-dimensional as possible. In particular, let $M_s$ be the $n\times n$ matrix whose $i^{th}$ row vector is $p_i^s - p_{0}^s$. Then clearly the rank of $M_s$ is equal to the dimension of the space spanned by $p_0^s,...,p_{n}^s$. We now want to add a regularization term to penalize high-rank matrices $M_s$. Unfortunately, no efficient methods exist for directly minimizing the rank of a matrix $M$. The relaxation suggested in \cite{fazel2002matrix} is to minimize the nuclear norm $\|M\|_*$ instead (i.e.\ the sum of the singular values of $M$). This technique was empirically shown to lead to low-rank matrix solutions in many applications, and is known to be equivalent to rank minimization in certain cases \cite{recht2010guaranteed}. The regularization term associated with $J_{\textit{type}}$ is thus given by
$$
J_{\textit{reg}}^1 = \sum_{s\in S} \| M_s\|_*
$$
To implement nuclear norm regularization, we have used the recently proposed method from \cite{hsieh2014nuclear}. 

We will also consider a variant in which the points $p_0^s,...,p_n^s$ are additionally required to be close to each other:
\begin{align*}
J_{\textit{type}}^{\textit{comb}} = \sum_{s\in S} \big(&\sum_{e\in E_s} \|p_{e} - \sum_{j=0}^{n} \lambda_{j}^{e,s} p_j^s\|^2 + \sum_{j=0}^n d(p_j^s,c_j^s) \big)
\end{align*}
where $c_j^s = \frac{1}{n+1}\sum_j p_j^s$ is the center-of-gravity of the points $p_0^s,...,p_n^s$. 

\subsection{Modelling relations}
Often we have information about how entities of different types are related, e.g.\ the fact that Steven Spielberg is the director of Jurassic Park. Such relationships can help us to align the subspaces corresponding to different types. 
Since our main aim is to improve the entity embeddings, rather than predicting relationships between entities of different types, methods such as TransH and TransR, which rely on projecting the entities to a different space, are not directly suitable. On the other hand, TransE is only suitable for one-to-one relations. 

We propose an alternative to TransE which is inspired by our modeling of semantic types. As in TransE, we assume that every relation $k$ is represented as a vector $r_k$.  We furthermore write $\textit{rhs}(e,k) = \{f \,|\, (e,k,f) \in G \}$ and $\textit{lhs}(k,f) = \{e \,|\, (e,k,f) \in G \}$. Rather than imposing that $ e + r_k = f$ if $(e,k,f)\in G$, as in TransE, we require that the points in $P_{e,k} = \{p_f \,|\, f\in \textit{rhs}(e,k)\}\cup \{p_e + r_k\}$ lie in a low-dimensional subspace and, similarly, that the points in $P_{k,f} = \{p_e \,|\, e\in \textit{lhs}(k,f) \} \cup \{p_f - r_k\}$ lie in a low-dimensional subspace. Note that in the case of one-to-many or many-to-one relations, this part of the model is similar to TransH in the special case where the considered subspaces are one-dimensional. Note that in the case of a one-to-many or many-to-many relation, the set of entities $\textit{rhs}(e,k)$ is essentially treated as an additional semantic type (e.g.\ the set of all films directed by Stephen Spielberg), and similar for many-to-one relations and the set $\textit{lhs}(k,f)$.
As for the semantic types we will consider a number of variants:
\begin{align*}
J_{\textit{rel}}^{\textit{dim}} &= \sum_{k \in R}  \sum_{p \in P_{e,k}} \|p - \sum_{j=0}^{n} \mu_{j}^{e,k} q_j^{e,k}\|^2 \\
&\quad\quad + \sum_{p \in P_{k,f}} \|p - \sum_{j=0}^{n} \mu_{j}^{k,f} q_j^{k,f}\|^2 \\
J_{\textit{rel}}^{\textit{dist}} &= \sum_{f\in \textit{rhs}(e,k)} d(p_f,p_e + r_k)^2 + \sum_{e\in \textit{rhs}(k,f)} d(p_e,p_f - r_k)^2\\
J_{\textit{rel}} &= J_{\textit{rel}} + J_{\textit{rel}}^{\textit{dist}}
\end{align*}
where we write e.g.\ $p \in P_{e,k}$  to sum over all entities $e$ and all points $p$ in $P_{e,k}$. Note that the variant $J_{\textit{rel}}^{\textit{dist}}$ essentially corresponds to TransE. For the variants $J_{\textit{rel}}$ and $J_{\textit{rel}}^{\textit{comb}}$ we again use nuclear norm regularization to enforce low-dimensional subspaces. Let the $i^{\textit{th}}$ row vector of the matrix $M_{e,k}$ be given by $q_i^{e,k} - q_0^{e,k} $ and similar for $M_{k,f}$. We define:
$$
J_{\textit{reg}}^2 = \sum_{k\in R} \| M_{e,k}\|_* + \| M_{k,f}\|_*
$$
Note that we only need to consider the combination $(e,k)$ or the combination $(k,f)$ if there is at least one triple of the form $(e,k,f)$ in $G$, since otherwise we can trivially choose $M_{e,k}$ and $M_{e,k}$ as the zero matrix. The full regularization term is given by $J_{\textit{reg}} = J_{\textit{reg}}^1 + J_{\textit{reg}}^2$.

\section{EVALUATION}\label{secEvaluation}

\subsection{Data acquisition}
In our experiments, we have used Wikidata to obtain a set of entities $E$ and their corresponding semantic types. To generate the bag-of-words representation $W_e$ of a given entity, we take advantage of the fact that Wikidata entities $e$ are linked to their corresponding Wikipedia article $d_e$. The set $W_e$ contains the words occurring in $d_e$, as well as the $m$ words before and after any mentions of the entity in other Wikipedia articles. Following \cite{glove2014}, we have used a window size of $m=10$ (but without crossing sentence boundaries). In particular, we treat every link from some Wikipedia article $d_x$ to $d_e$ as a mention of $e$, as well as any repeated occurrences of the corresponding anchor text in $d_x$. The word-word co-occurrence in the $J_{\textit{glove}}$ component of our model has been obtained from the entire Wikipedia corpus, as in the standard GloVe model. Using the Wikidata dump from October 26, 2015 and the Wikipedia dump from November 02, 2015, we have then selected those entities $e$ which are mentioned in at least 10 Wikipedia articles, resulting in a set $E$ containing 1,292,702 entities. For each semantic type $s$, the set $E_s$ contains those entities which are asserted to be of type $s$ via the \textit{instance of} property as well as all instances which are asserted to belong to one of the supertypes of $s$, which was determined using the \textit{subclass of} property. As the set of binary relations $R$ we considered all Wikidata properties whose value is another entity, apart from \textit{instance of} and \textit{subclass of} which have already been used to determine the sets $E_s$. In the case of Wikipedia, we adopted a fairly straightforward preprocessing strategy, as used in many other works such as \cite{wang2014knowledge}. In particular, we removed punctuations, lower-cased the tokens, and conducted sentence segmentation using the NLTK library\footnote{\url{http://www.nltk.org/}}. We also removed words whose term frequency in the entire collection was less than 10. A script has been made available online\footnote{\url{https://github.com/bashthebuilder/ECAI-2016/blob/master/README.md}}, which generates an exact copy of our data set, starting from the publicly available dumps of Wikipedia and Wikidata. The implementation of all variants of our model has also been made available at the same link.

Most knowledge graph embedding models have been evaluated on fragments of Freebase and WordNet. Our choice of Wikidata is motivated by the fact that it has relatively clean semantic type information. For example, while Barack Obama is of type \textit{Human} on Wikidata, Freebase among others mentions the following types: \textit{film subject}, \textit{musical artist} and \textit{building occupant}. Furthermore, while Freebase contains information about tens of millions of entities, the standard benchmark datasets, called FB15k \cite{NIPS2013_5071} and FB13 \cite{socher2013reasoning}, are relatively small: FB15k covers 14,951 entities and 1,345 relation types, while FB13 covers 74,043 entities and 13 relation types. For completeness, we will include a comparison of our model on these standard benchmark sets for link prediction and triple classification, which are the two standard evaluation tasks for knowledge graph embedding. Our other experiments will be oriented more towards evaluating the usefulness of our model for learning conceptual space representations, in particular their ability to capture semantic relations between entities of the same type (which are not covered in the knowledge graph). This requires a sufficient number of entities for each of the considered semantic types, and a sufficiently clean semantic type structure. Accordingly these tasks will be evaluated only on the WikiData fragment described above. Finally, WordNet has a rich semantic type hierarchy, but contains relatively few instances of these types (e.g.\ of the 51K leaf nodes in WordNet 1.7 only 7K were found to be instances in \cite{alfonseca2002distinguishing}) and is thus not suitable for our purposes.   

The semantic types of the entities occurring in FB15k and FB13 have been obtained from the ``type/instance'' field in the Freebase dump\footnote{\url{https://developers.google.com/freebase/data}}. To link Freebase entities to Wikipedia, we have made use of existing Freebase-WikiData mappings\footnote{\url{https://developers.google.com/freebase/data#freebase-wikidata-mappings}}.

\subsection{Variants and baseline methods}
Our main baseline is pTransE, which also learns an embedding of entities by combining a word embedding model with a knowledge graph embedding model. We used the it's publicly available implementation\footnote{https://github.com/Mrlyk423/Relation\_Extraction}. We consider three variants of this baseline: pTransE$_{\text{anch}}$ is the version proposed in \cite{wang2014knowledge}, which uses anchor text for aligning word vectors and entity vectors; pTransE$_{\text{art}}$ is the improvement proposed in \cite{DBLP:conf/emnlp/ZhongZWWC15}, which uses the words in the Wikipedia article $d_e$ instead of anchor text (and a slightly different model); pTransE$_{\text{full}}$ is a variant of pTransE$_{\text{art}}$, which uses the bag of words representation $W_e$ instead, as in our method. In addition, we compare our method against RESCAL, as well as a number of knowledge graph embedding methods: TransE, TransH, TransR and CTransR. The source codes of these translation-based models are publicly available online\footnote{https://github.com/Mrlyk423/Relation\_Extraction}. RESCAL \cite{nickel2011three} is a collective matrix factorization model based on tensor factorization, which has been designed to account for the inherent structure of dyadic relational data. The implementation of RESCAL can be found here\footnote{https://github.com/mnick/rescal.py}. For the knowledge graph embedding methods, we used Bernoulli sampling for selecting negative examples (see \cite{wang2014knowledge}); we also obtained results for uniform sampling (not shown), and found the results to be very similar to Bernoulli sampling but slightly worse. It is expected that all of these methods will perform worse than both pTransE and our model, as they cannot exploit the text representation $W_e$ of the entities. We also compare our method with Skip-gram and CBOW, which can only use text representations and are thus also expected to perform worse. In particular, to apply these models to learn entity embeddings, we use the same method as for our model to determine entity mentions on Wikipedia, and then apply the standard models based on the words surrounding these mentions. Finally, we have compared our method with the multi-dimensional scaling (MDS) based method from \cite{derracAIJ}, in which case we learn a separate vector space for every semantic type. Because of the limited scalability of the latter model, however, we have only considered this for semantic types with up to 10000 instances. Following \cite{derracAIJ}, for each of the remaining semantic types, a vector space representation of the corresponding entities was obtained using Positive Pointwise Mutual Information (PPMI). We then applied multi-dimensional scaling to obtain a lower-dimensional representation, using the angular difference between the initial vectors as metric. We have used the MDS model implemented in MATLAB. We have also considered the method from \cite{DBLP:conf/acl/GuoWWWG15} as an additional baseline, but found that this method could not scale to even the reduced data set that we used for the MDS experiments.

\begin{table}[t]
\begin{center}
{\caption{Overview of considered variants of our model. \label{tabOverviewVariants}}}
\footnotesize
\begin{tabular}{|l|ccc|}
\hline
name & type & relation & regularization\\
\hline
EECS$_{\text{full}}$    & $J_{\textit{type}}$ & $J_{\textit{rel}}$ & $J_{\textit{reg}}^1 + J_{\textit{reg}}^2$\\
EECS$_{\text{no rel}}$  & $J_{\textit{type}}$ & - & $J_{\textit{reg}}^1$\\
EECS$_{\text{no type}}$ & - & $J_{\textit{rel}}$ & $J_{\textit{reg}}^2$\\
EECS$_{\text{no NN}}$    & - & $J_{\textit{rel}}$ & -\\
EECS$_{\text{text}}$     & - & - & -\\
\hline
EECS$_{\text{rel-dim}}$   & $J_{\textit{type}}$ & $J_{\textit{rel}}^{\textit{dim}}$ & $J_{\textit{reg}}^1 + J_{\textit{reg}}^2$\\
EECS$_{\text{rel-dist}}$  & $J_{\textit{type}}$ & $J_{\textit{rel}}^{\textit{dist}}$ & $J_{\textit{reg}}^1$\\
EECS$_{\text{type-comb}}$ & $J_{\textit{type}}^{\textit{comb}}$ & $J_{\textit{rel}}$ & $J_{\textit{reg}}^1 + J_{\textit{reg}}^2$\\
EECS$_{\text{type-dist}}$ & $J_{\textit{type}}^{\textit{comb}}$ & $J_{\textit{rel}}$ & $J_{\textit{reg}}^2$\\
\hline
\end{tabular}
\end{center}
\end{table}

Throughout this section, we will refer to our model as EECS (Entity Embeddings with Conceptual Subspaces).
As an ablation study, we will consider a number of variants of our model in which some components have been removed.  EECS$_{\text{full}}$ refers to our full model, in which $J_{\textit{type}}$ is used for modelling semantic types and $J_{\textit{rel}}$ is used for modelling relations; EECS$_{\text{no rel}}$ refers to a variant in which $J_{\textit{rel}}$ and the associated regularization component $J_{\textit{reg}}^2$ have been removed; EECS$_{\text{no type}}$ refers to a variant in which $J_{\textit{type}}$ and $J_{\textit{reg}}^1$ have been removed;  EECS$_{\text{no NN}}$ refers to a variant in which the regularization component $J_{\textit{reg}}$ has been removed (which also trivializes the component $J_{\textit{type}}$); EECS$_{\text{text}}$ refers to a variant in which only the component $J_{\textit{text}}$ is used, reducing our model essentially to a variant of GloVe.

Furthermore, we have considered a few variants of EECS$_{\text{full}}$ in which we change the component $J_{\textit{type}}$ or $J_{\textit{rel}}$ by one of the proposed alternatives: EECS$_{\text{rel-dim}}$ refers to a variant in which $J_{\textit{rel}}$ is replaced by $J_{\textit{rel}}^{\textit{dim}}$, EECS$_{\text{rel-dist}}$ refers to a variant in which $J_{\textit{rel}}$ is replaced by $J_{\textit{rel}}^{\textit{dist}}$, EECS$_{\text{type-comb}}$ refers to a variant in which $J_{\textit{type}}$ has been replaced by $J_{\textit{type}}^{\textit{comb}}$, and EECS$_{\text{type-dist}}$ refers to a variant in which only distance information is considered for modelling semantic types (which corresponds to using $J_{\textit{type}}^{\textit{comb}}$ without regularization).
An overview of the considered variants of our model is provided in Table \ref{tabOverviewVariants}.

\subsection{Methodology}
All experiments were evaluated using five-fold cross validation. For tuning the parameter \(\beta\) of our model, based on a tuning/validation set in each experiment, we considered the range $\{50,100,150,200,250,300,350,400\}$. For the parameter \(\alpha\), we considered values between 0 and 1 with an increment of 0.1. The number of iterations for all models was set to 20, as we found that beyond this number empirical results became fairly consistent in all cases. Based on the tuning set, in each of the experiments the optimal value of \(\beta\) was found to be 300, while the optimal values of $\alpha$ varied between 0.4 and 0.7. The number of dimensions was always set to 300 for our model, noting that because of the nuclear norm regularization this only represents an upper bound on the actual number of dimensions.  All parameters of the baseline methods, including the number of dimensions, have been optimized based on the tuning set in each experiment. For the MDS method, the number of dimensions was tuned for each semantic type separately (as this method learns a separate vector space for each semantic type), considering the range from 10 to 100 in steps of 10. For the remaining baselines, which construct a single vector space, the number of dimensions was varied between 50 to 300 in steps of 50.

Our model has been implemented in C using standard POSIX threads, which helps scale our implementation to large text collections. For example, for the considered 1.2 million Wikidata entities our full model takes about 30 minutes per iteration using 8 threads, scaling almost linearly in the number of entities. In contrast, EECS$_{\text{text}}$ takes about 18 minutes for each iteration using 8 threads.

\subsection{Results}

We will evaluate our model on four different tasks: \textit{ranking}, \textit{induction}, \textit{analogy making}, and \textit{knowledge graph embedding}. The first two of these tasks are directly aimed at evaluating to what extent the type-specific subspaces learned by our model are useful as conceptual space representations. In particular, \textit{ranking} will evaluate to what extent important features of a given semantic type can indeed be modelled as directions in the associated subspace, while \textit{induction} assesses to what extent we can use these representations to find new instances of a given concept, given only a few example instances. The \textit{analogy making} task is aimed at evaluating how well the different subspaces are aligned. As discussed above, these first three tasks will be evaluated using a large fragment of WikiData. 
The motivation behind the fourth tasks relates to the observation that even though our motivation was rather different from the motivation behind pTransE, both our model and pTransE combine a word based entity embedding component with a knowledge graph embedding component. Since pTransE has proven a successful approach for knowledge graph embedding, we want to analyse whether our model has any advantages in such a setting. As explained above, for this task we will use the standard benchmark datasets FB15k and FB13.

\begin{table}[t]
\begin{center}
\caption{Number of dimensions selected by the nuclear norm (NN) regularization component of our model for some of the semantic types.}
\footnotesize
\begin{tabular}{|c|c|c|}
\hline
Semantic Type & Number of Entities & NN-Dimensions \\ \hline
  human            &   191211   &  288               \\ 
      railway station & 4120 & 121 \\ 
    house          &        2762            &        136       \\ 
      organization        &     1379             &  88           \\ 
      national park & 1307 & 56 \\ 
      building & 1269 & 52 \\ 
      food & 1155 & 55 \\ 
college & 858 & 33 \\ 
automobile    & 31                & 12            \\ 
      candy & 10 & 2 \\ \hline
\end{tabular}
\label{nn-dimensions-table}
\end{center}
\end{table}

An important aspect in the discussion of the result is to analyze the effectiveness of nuclear norm regularization in identifying the most appropriate number of dimensions for each of the semantic types. To illustrate the behaviour of this regularization component, Table \ref{nn-dimensions-table} shows the number of dimensions that was found for a few notable semantic types (when using the default configuration of our model). As expected, semantic types with more entities generally end up being associated with higher-dimensional subspaces, but other factors affect the choice as well. For example, note that \textit{house} has fewer instances than \textit{railway station}, while being represented by a higher-dimensional subspace. Intuitively, this reflects the idea that the type \textit{house} is more diverse or complex than the type \textit{railway station}. 

\begin{table}[t]
    \begin{center}
    \caption{Experimental results for the full WikiData test data.}
    \footnotesize
    \begin{tabular}{|l|c|ccc|c|}
    \hline
         & Ranking  & \multicolumn{3}{|c|}{Induction} & Analogy \\
         & $\rho$  & MAP & P@5 & MRR & Acc.\ \\
    \hline     
    Skip-gram   & 0.155 & 0.176 & 0.356 & 0.505 & 0.184 \\ 
    CBOW        & 0.159 & 0.182 & 0.350 & 0.500 & 0.213 \\ 
    \hline
    RESCAL & 0.081 & 0.020 & 0.189 & 0.423 & 0.371 \\
    \hline
    TransE        & 0.110 & 0.060 & 0.200 & 0.451 & 0.382 \\ 
    TransH        & 0.142 & 0.072 & 0.210 & 0.415 & 0.382 \\ 
    TransR        & 0.100 & 0.102 & 0.302 & 0.489 & 0.378 \\ 
    CTransR       & 0.122 & 0.132 & 0.323 & 0.499  & 0.402 \\ 
    \hline
    pTransE$_\text{anch}$     & 0.099 & 0.101 & 0.301 & 0.488 & 0.476\\ 
    pTransE$_\text{art}$    & 0.202 & 0.218 & 0.475 & 0.751 & 0.512 \\ 
    pTransE$_\text{full}$       & 0.213 & 0.224 & 0.490 & 0.756 & 0.532\\ 
    \hline
    EECS$_{\text{full}}$        & \textbf{0.319} & \textbf{0.231} & \textbf{0.609} & \textbf{0.883} & 0.591\\ 
    EECS$_{\text{no rel}}$      & 0.301 & 0.229 & 0.588 & 0.868 & 0.552\\ 
    EECS$_{\text{no type}}$     & 0.266 & 0.225 & 0.585 & 0.854 & 0.549\\ 
    EECS$_{\text{no NN}}$       & 0.258 & 0.220 & 0.581 & 0.843 & 0.545\\ 
    EECS$_{\text{text}}$        & 0.254 & 0.218 & 0.579 & 0.831 & 0.540\\
    \hline 
    EECS$_{\text{type-comb}}$   & 0.312 & \textbf{0.231} & 0.601 & \textbf{0.883} & \textbf{0.595}\\ 
    EECS$_{\text{type-dist}}$ & 0.295 & \textbf{0.231} & 0.585 & 0.858 & 0.550\\ 
    EECS$_{\text{rel-dim}}$     & 0.309 & 0.225 & 0.585 & 0.859 & 0.551\\
    EECS$_{\text{rel-dist}}$ & 0.299 & 0.225 & 0.585 & 0.855 & 0.549 \\ 
    \hline
    \end{tabular}
    \label{tabResults}
    \end{center}
%
    \begin{center}
    \caption{Comparison with MDS on a subset of the WikiData test data.}
    \footnotesize
    \begin{tabular}{|l|c|ccc|c|}
    \hline
         & Ranking  & \multicolumn{3}{|c|}{Induction} & Analogy \\
         & $\rho$  & MAP & P@5 & MRR & Acc.\ \\
    \hline     
    MDS                         & 0.101 &  0.121 & 0.231 & 0.388 & 0.354 \\ 
    EECS$_{\text{full}}$        & 0.218 & 0.140 & 0.301 & 0.463 & 0.456 \\ 
    \hline
    \end{tabular}
    \label{tabResMDS}
    \end{center}
\end{table}

\subsubsection{Ranking}
A characteristic feature of conceptual spaces is that they are encoded as Cartesian products of interpretable dimensions. For a vector space model to be meaningful as a conceptual space, it is therefore important that salient properties can be modelled as directions\footnote{Conceptual space representations also encode information about the correlation between the underlying dimensions, which in our case is captured by the angles between these directions.}. Therefore, we have evaluated the ability of our model to correctly rank entities according to a given property. As we need the ground truth, we have focused on properties with numerical values which are available in Wikidata (but have not been considered when learning the space), e.g.\ the \textit{date of birth} for entities of type \textit{human}, or the \textit{boiling point} of entities of type \textit{chemical element}. In total, we have retrieved 26 numerical attributes which are available for at least 30 entities. Some of these numerical attributes appear for different semantic types (e.g.\ the property \textit{inception}, referring to the foundation year, applies to the semantic types \textit{film}, \textit{organization} and \textit{country}, among others). In total, we obtained 73 such property-type combinations, by considering for each numerical attribute all the maximally specific semantic types with at least 30 instances that have the attribute. Each of these 73 combinations was considered as a problem instance.
For each problem instance, the corresponding set of entities is split into 60\% training, 20\% validation and 20\% testing sets. 
The full specification of the 73 problem instances and corresponding splits is available online\footnote{https://github.com/bashthebuilder/ECAI-2016}.
From the training set, a direction is estimated using the SVMRank model\footnote{\url{https://www.cs.cornell.edu/people/tj/svm_light/svm_rank.html}} \cite{joachims2002optimizing}. The parameters of the resulting ranking models are optimized using the validation sets. Table \ref{tabResults} shows the performance on the testing set, in terms of Spearman's $\rho$\footnote{The reported average $\rho$ values have been obtained using the Fisher z-transformation.}, expressing the correlation between the ranking predicted by the model and the ranking according to the numerical values found in Wikidata.

The results show that standard word and knowledge graph embedding models are not competitive, which is not surprising given that they use less information than our model. However, the results also show that our model substantially outperforms pTransE, even the variant pTransE$_{\text{full}}$ which uses the same input as our model. Comparing the results for the variants of our model, we notice that the relation component only has a small impact, i.e.\ the performance of EECS$_{\text{no rel}}$ is close to EECS$_{\text{full}}$, and similarly, the performance of EECS$_{\text{no type}}$ is close to EECS$_{\text{text}}$. Regarding the variants of $J_{\textit{type}}$, EECS$_{\text{full}}$ and EECS$_{\text{type-comb}}$ are clearly better than EECS$_{\text{type-dist}}$, which shows the importance of nuclear norm regularization for identifying low-dimensional subspaces.
The results in Table \ref{tabResMDS} compare our model against the MDS model from \cite{derracAIJ} on a reduced set of 27 problem instances. These results clearly show that the MDS method is not competitive.


\begin{table}[t]
\begin{center}
\caption{Five lowest ranked entities for a number of ranking problem instances.}
\label{enity_ranking_top}
\footnotesize
\begin{tabular}{|lll|}
\hline
Population & Inception & Date of Birth  \\
\hline
Malta            &     General Electric          &      Valmiki                         \\
Bermuda            &      IBM         &             Jesus Christ            \\
 Monaco            &        Hewlett Packard       &      Cleopatra                         \\
San Marino              &   Microsoft            &     Ptolemy                         \\
Barbados              &      Oracle Corporation         &   Plato                            \\
\hline
\end{tabular}
\end{center}
%
\begin{center}
\caption{Five highest ranked entities for a number of ranking problem instances.}
\label{enity_ranking_bottom}
\footnotesize
\begin{tabular}{|lll|} \hline
Population & Inception & Date of Birth  \\
\hline
  China          &    Alphabet Inc.           &     Prince George of Cambridge                          \\
India        &      Tencent Holdings         &          Isabela Moner                     \\
USA             &   Facebook, Inc.            &    Justin Bieber                           \\
Soviet Union              &      Uber         &        Lionel Messi                       \\
Brazil              &      Amazon.com         &      Kim Kardashian                        \\
\hline
\end{tabular}
\end{center}
\end{table}

Tables \ref{enity_ranking_top} and \ref{enity_ranking_bottom} illustrate the results of the ranking experiment for three attributes, by showing the 5 lowest and 5 highest ranked entities respectively. Note that Table \ref{enity_ranking_top} starts with the lowest ranked entity (i.e.\ the entity that has the lowest value for the considered attribute), while Table \ref{enity_ranking_bottom} starts with the highest ranked entity (i.e.\ the entity that has the highest value for the considered attribute). While the rankings are not perfect (e.g.\ Bermuda, Monaco, San Marino and Barbados are all less populous than Malta, Ptolemy lived around 500 years after Plato), the model's ability to separate high-scoring entities from low-scoring entities is nonetheless remarkable, considering that none of the information that was used to learn the vector space explicitly referred to these attributes.

\subsubsection{Induction}
A second characteristic feature of conceptual spaces is that properties correspond to convex regions. Moreover, it is often assumed that the boundaries of these regions are determined based on the distance to a particular point in the space, which acts as a prototype. In this experiment, we test our method's ability to make inductive inferences based on this view. In particular, given a number of entities of the same type which have some property in common, the task we consider is to identify other entities that also have this property (without any knowledge about the property being considered). 

Problem instances in this case were obtained by omitting all triples of the form $(.,r,f)$ for particular choices of $r$ and $f$, when learning the embeddings. The set of entities $e$ for which $(e,r,f)\in G$ then defines a problem instance. For example, the property being considered could be ``films directed by Stephen Spielberg''. Given a few examples of such films, the task is to identify other films directed by Stephen Spielberg (but without the knowledge that this is the property being considered). For each $(r,f)$ combination, the set of entities $\{e \,|\, (e,r,f)\in G\}$ is split into 60\% training, 20\% tuning and 20\% testing sets. Details on the $(r,f)$ combinations and associated splits are available online\footnote{\url{https://github.com/bashthebuilder/ECAI-2016}}.

For evaluation purposes, we consider this task as a ranking task. In particular, for each problem instance, we rank the entities of the associated semantic type (defined as the most specific semantic type that contains all the considered entities) based on their distance to the center-of-gravity of the training instances, and evaluate the quality of this ranking using mean average precision (MAP), Precision@5 (P@5) and Mean Reciprocal Rank (MRR); note that in all cases, higher values are better.

The results in Table \ref{tabResults} show that our model again substantially outperforms all of the baselines. Note, however, that in the case of MAP, the differences with pTransE$_{\text{full}}$ are rather small. The fact that the differences are clearer for P@5 and MRR suggests that our method is better able to select a few entities with high precision. The MAP score tends to be dominated by outliers, leading to smaller differences. Regarding the different variants of our model, the semantic type component and relation component now play a more equal role, given the rather similar performance of EECS$_{\text{no rel}}$ and EECS$_{\text{no type}}$, although semantic type information is still more important than the knowledge graph information (as EECS$_{\text{no rel}}$ performs better than EECS$_{\text{no type}}$). As for the ranking experiment, we notice that using $J_{\textit{type}}^{dist}$ in EECS$_{\text{no type}}$ leads to worse results, highlighting again the importance of nuclear norm regularization. A before, the MDS model is not competitive.



\subsubsection{Analogy making}
Finally we have considered the problem of completing analogical proportions of the form ``$a$ is to $b$ what $c$ is to ...'', which is a standard evaluation task for word embeddings. Our main aim in this task is to evaluate how well different subspaces are aligned. We have used the test sets from the GloVe project\footnote{\url{http://nlp.stanford.edu/projects/glove/}} that are about entities, resulting in a total of 8363 problem instances. As there is no need for training data, in this case we randomly split the data into 25\% tuning and 75\% testing sets. Full details on the test sets and splits that were used have been made available online\footnote{\url{https://github.com/bashthebuilder/ECAI-2016}}.

The results are largely consistent with the findings from the previous two experiments. The main difference is that the variant EECS$_{\text{type-comb}}$ slightly outperforms EECS$_{\text{full}}$ in this case. The rather large difference in performance between EECS$_{\text{type-comb}}$ and EECS$_{\text{type-dist}}$ again clearly illustrates the impact of nuclear norm regularization on the results.

\subsubsection{Knowledge graph embedding}
We have also conducted two knowledge graph embedding experiments using the benchmark datasets FB15k and FB13. In particular, we have evaluated our method on the widely used Link Prediction and Triple Classification tasks. 

\paragraph{Link prediction} For the link prediction task \cite{NIPS2013_5071}, given an entity $e$ and a relation $r$, the aim is either to find an entity $f$ such that $(e,r,f)$ or to find an entity $f$ such that $(f,r,e)$. We have used the standard FB15k test set for this evaluation, which allows us to compare our model with the published results of the state-of-the-art knowledge graph embedding models. Two widely used evaluation metrics, which we will also use, are the average rank of correct entities, called ``Mean Rank'', and ``HITS@10'', which is defined as the proportion of test triples in which the target entity was ranked in the top 10. Note that the Mean Rank score is to be minimized while the HITS@10 score is to be maximized. We have used the standard evaluation protocol, including the Bernoulli sampling trick to corrupt the head or tail entity. Test instances were not filtered (which corresponds to the so-called raw version of the task).

In Table~\ref{link_prediction_triple_classification_results} we show that our model clearly outperforms the standard baselines in both metrics. To a large extent, this is due to the fact that, apart from pTransE, the other models do not exploit the bag-of-words representations of the entities.
Note that we do not show results for EECS$_{\text{text}}$ and EECS$_{\text{no rel}}$ in the tables because these models do not take into account any input from the knowledge graph, and is therefore not suitable. The baselines Skip-gram and CBOW are not considered for the same reason.

\begin{table}
\centering
\caption{Link prediction and Triple classification results.}
\label{link_prediction_triple_classification_results}
\footnotesize
\begin{tabular}{|l|cc|cc|}
\hline
Models                    & \multicolumn{2}{|c|}{Link Prediction (FB15k)} & \multicolumn{2}{c|}{Triple Classification} \\ \hline
                          & Mean Rank         & HITS@10         & FB13                & FB15k               \\ \hline
RESCAL                    &       683            &      44.1           &   65.3     & 71.6                \\ \hline
TransE                    & 125 & 47.1                 &     81.5   &       79.8              \\
TransH                    & 87  & 64.4                 &        \textbf{83.3}     & 79.9            \\
TransR                    & 77  & 68.7                 &    82.5                 &  82.1           \\
CTransR                  & 75  & 70.2                 &         -            &      84.3           \\ \hline
pTransE$_\text{anch}$     & 58 & 84.6                 &         73.3            &   74.3                  \\
pTransE$_\text{art}$      & 55 & 85.3                 &         75.8            &   75.5                  \\
pTransE$_\text{full}$     & 51 & 86.4                &          76.3        &       77.4          \\ \hline
EECS$_{\text{full}}$      & 48   & 89.7                  &      83.1               &    89.6     \\
EECS$_{\text{no type}}$   & 56 & 84.7                 &         71.2            &       82.1      \\
EECS$_{\text{no NN}}$     & 59   & 82.7                &        70.1         &          81.4    \\ \hline
EECS$_{\text{type-comb}}$ & \textbf{47}   & \textbf{89.9}        & \textbf{83.3}          & \textbf{89.9}    \\
EECS$_{\text{type-dist}}$ & 54   & 83.2                 &           81.1          &         82.1    \\
EECS$_{\text{rel-dim}}$   & 54   & 85.1                 &       79.3              &     88.2\\
EECS$_{\text{rel-dist}}$  & 52   & 85.3                 &           78.8          &     87.1 \\
\hline              
\end{tabular}
\end{table}

\paragraph{Triple classification} The objective in triple classification \cite{socher2013reasoning} is to judge whether a given triplet $(e,r,f)$ is correct or not, i.e.\ whether entities $e$ and $f$ are in relation $r$ with each other. This can be naturally cast as a binary classification task. We present results on FB13 and FB15k. The FB13 dataset already comes with golden negative triplets, while we followed the methodology from \cite{socher2013reasoning} to construct negative samples for FB15k. For the classification task, we need to set a threshold \(\delta_r\) for each relation \(r\). We obtain \(\delta_r\) by maximizing the classification accuracies on the validation set. For the given triplet, if the energy score is larger than the relation-specific \(\delta_r\), the instance will be classified as positive, otherwise negative. This is the standard experimental setting for this evaluation task.

Our experimental results are shown in Table~\ref{link_prediction_triple_classification_results}. With the exception of the pTransE variants, we again show the published results for the baseline models. The baseline results have been reported in \cite{xiao2015transg, he2015learning, ji2015knowledge, nguyen2016stranse}. On the FB13 dataset, our model matches the performance of the TransH model, although we clearly outperform all baselines on the FB15k dataset. This means that on the FB13 dataset, the bag-of-words representations of the entities cannot be exploited effectively, although our model is still not at a disadvantage. This seems related to the fact that only 13 relation types are considered in FB13, which were moreover specifically selected such that they can be predicted from each other, in the sense that hard-to-predict relation types have been removed \cite{socher2013reasoning}.

\section{CONCLUSIONS}
We have proposed a new method for learning vector-space embeddings of entities, based on available semantic information (from Wikidata) and textual descriptions (from Wikipedia). From a technical point of view, the main novelty of our model is the use of nuclear norm regularization to encode the requirement that entities of the same semantic type should lie in a lower-dimensional subspace. In particular, nuclear norm regularization allows the model to automatically select the most appropriate number of dimensions for the subspace corresponding to each type. 
From an application point of view, our main motivation was to learn subspaces that are useful as approximations of conceptual spaces. To support this view, among others, we have shown that many numerical attributes can be faithfully modelled as directions and that the learned representations allow us to model induction based on distance to a centroid. In addition, we have also obtained good results for analogy making and for two standard knowledge graph embedding tasks.

\section{ACKNOWLEDGEMENTS}
This work was supported by ERC Starting Grant 637277.

\bibliographystyle{ecai}
\bibliography{commonsense}

\end{document}